\documentclass[11pt,a4paper]{article}
\usepackage[hyperref]{acl2021}
\usepackage{times}
\usepackage{latexsym}

\usepackage{xcolor}

\usepackage{microtype}

\usepackage{url}

\usepackage{graphicx} 
\usepackage{subcaption} 
\usepackage{comment} 
\usepackage{makecell} 
\usepackage{gb4e} 
\noautomath 

\aclfinalcopy 
\def\aclpaperid{3043} 

\title{Assessing the Syntactic Capabilities of Transformer-based Multilingual Language Models}

\author{
  Laura Pérez-Mayos\textsuperscript{1}, Alba Táboas García\textsuperscript{1}, Simon Mille\textsuperscript{1}, Leo Wanner\textsuperscript{2,1} \\[5pt]
  \textsuperscript{1} TALN Research Group, Pompeu Fabra University, Barcelona, Spain \\
  \textsuperscript{2} Catalan Institute for Research and Advanced Studies (ICREA), Barcelona, Spain \\[5pt]
  \texttt{\{laura.perezm$\mid$alba.taboas$\mid$simon.mille$\mid$leo.wanner\}@upf.edu}
}

\date{}

\begin{document}
\maketitle

\begin{abstract}
Multilingual Transformer-based language models, usually pretrained on more than 100 languages, have been shown to achieve outstanding results in a wide range of cross-lingual transfer tasks. However, it remains unknown whether the optimization for different languages conditions the capacity  of the models to generalize over syntactic structures, and how languages with syntactic phenomena of different complexity are affected. In this work, we explore the syntactic generalization capabilities of the monolingual and multilingual versions of BERT and RoBERTa.
More specifically, we evaluate the syntactic generalization potential of the models on English and Spanish tests, comparing the syntactic abilities of monolingual and multilingual models on the same language (English), and of multilingual models on two different languages (English and Spanish). For English, we use the available SyntaxGym test suite; for Spanish, we introduce SyntaxGymES, a novel ensemble of targeted syntactic tests in Spanish, designed to evaluate the syntactic generalization capabilities of language models through the SyntaxGym online platform.
\end{abstract}

\section{Introduction}
\label{sec:introduction}

Transformer-based neural models such as BERT \cite{devlin2018bert}, RoBERTa \cite{liu2019roberta}, DistilBERT \cite{Sanh2019DistilBERTAD}, XLNet \cite{NEURIPS2019_dc6a7e65}, etc. are excellent learners. They have been shown to capture a range of different types of linguistic information, from morphological \cite{Edmiston2020} over syntactic \cite{hewitt-manning-2019-structural} to lexico-semantic \cite{joshi2020contextualized}. A particularly significant number of works study the degree to which these models capture and generalize over (i.e., learn to instantiate correctly in different contexts) syntactic phenomena, including, e.g., subject-verb agreement, long distance dependencies, garden path constructions, etc. \cite{linzen2016assessing,marvin-linzen-2018-targeted,futrell-etal-2019-neural,wilcox2019hierarchical}. However, most of these works focus on monolingual models, and, if the coverage of syntactic phenomena is considered systematically and in detail, it is mainly for English, as, e.g., \cite{hu2020systematic}. This paper aims to shift the attention from monolingual to multilingual models and to emphasize the importance to also consider the syntactic phenomena of languages other than English when assessing the generalization potential of a model. More specifically, it systematically assesses how well multilingual models are capable to generalize over certain syntactic phenomena, compared to monolingual models, and how well they can do it not only for English, but also for Spanish.

Multilingual models such as mBERT (multilingual BERT,  \citep{devlin2018bert}), XLM \cite{lample2019crosslingual} and XLM-R \cite{conneau-etal-2020-unsupervised} proved to achieve outstanding performance on cross-lingual language understanding tasks, including on low-resource languages for which only little training data is available. However, these models face the risk of running into what \citet{conneau-etal-2020-unsupervised} refer to as ``curse of multilinguality": adding languages to the model increases the performance on low-resource languages up to a point, after which the overall performance on monolingual and cross-lingual benchmarks degrades. The question is thus whether, and if yes to what degree, this degradation affects the syntactic generalization potential of multilingual models across languages.

The reason to extend the evaluation to other languages (in our case, Spanish) is that many existing syntactic phenomena such as determiner and adjective agreement within the noun phrase, subject pro-drop, or flexible word order -- to name only a few -- are not prominent or do not exist in English, while in Spanish all of them do.


Our evaluation methodology is similar to that by \citet{hu2020systematic}, who test 20 model type combinations and data sizes on 34 English syntactic test suites, and find substantial differences in the syntactic generalization performance across different models. We draw upon their tests to test the syntactic generalization potential of monolingual and multilingual transformer-based models for English, and upon the Spanish SyntaxGym introduced in this paper for Spanish. Tu run the tests, we use the SyntaxGym toolkit \citep{gauthier2020syntaxgym}.

Our results show that, indeed, there is a substantial difference between the syntactic generalization potential of monolingual and multilingual models. But this difference depends on the language: While for English monolingual models (BERT and RoBERTa) offer a higher syntactic generalization than multilingual models (mBERT and XLM-R), this is not the case for Spanish, for which multilingual models (XLM-R) generalize better. Furthermore, multilingual models do not generalize equally well across languages, with mBERT generalizing, in general, better in English and XLM-R better in Spanish. Our experiments also show that it depends on the language how well a multilingual model captures a specific syntactic phenomenon such as, e.g., Agreement, Center-embedding or Garden Path.

The remainder of the paper is structured as follows. Section \ref{sec:related_work} introduces the work that is related to ours in terms of the evaluation methodology and, in particular, in terms of the assessment of multilingual language models. Section \ref{sec:test_suites} describes the English test suites, and presents the novel Spanish SyntaxGym test suites. Section \ref{sec:experiments} details the models that we tested and outlines how we use them to evaluate the probability of a text sequence. Section \ref{sec:discussion} offers a detailed analysis of the syntactic generalization abilities of the monolingual and multilingual versions of BERT and RoBERTa, and Section \ref{sec:conclusions} summarizes the implications that our work has for the use of multilingual language models.

\section{Related Work}
\label{sec:related_work}


Our work on the evaluation of the capability of monolingual and multilingual transformer-based LMs to capture syntactic information is in line with a number of previous works, including, e.g., those that are based on psycholinguistic experiments, focusing on highly specific measures of language modeling performance and allowing to distinguish models with human-like representations of syntactic structure \citep{linzen2016assessing,lau2017grammaticality,gulordava2018colorless,marvin-linzen-2018-targeted,futrell-etal-2019-neural}. Supervised probing models have been used to test for the presence of a wide range of linguistic phenomena \citep{conneau-etal-2018-cram, hewitt-manning-2019-structural, liu-etal-2019-linguistic, tenney2018what, voita2020information, elazar2020bert}. \citet{warstadt2020blimp}  isolate specific phenomena in syntax, morphology, and semantics, finding that state-of-the-art models struggle with some subtle semantic and syntactic phenomena, such as negative polarity items and extraction islands.

Recently, a number of works also address the cross-language assessment of models. \citet{pmlr-v119-hu20b} introduces XTREME, a multi-task benchmark for evaluating the cross-lingual generalization capabilities of multilingual representations across 40 languages and 9 tasks. They show that while XLM-R reduces the difference between the performance on the English test set and all other languages compared to mBERT for tasks such as XQuAD and MLQA, it does not have the same impact on structured prediction tasks such as PoS and NER. \citet{mueller2020cross} introduces a set of subject-verb agreement tests, showing that mBERT performs better than English BERT on Sentential Complements, Short VP Coordination, and Across a Prepositional Phrase, but worse on Within-an-Object Relative Clause, Across-an-Object Relative Clause and in Reflexive Anaphora Across a Relative Clause, and offers high syntactic accuracy on English, but noticeable deficiencies on other languages, most notably on those that do not use Latin script, as also noted by \citet{pmlr-v119-hu20b}. On the same line, \citet{ronnqvist-etal-2019-multilingual} concludes that mBERT is not able to substitute a well-trained monolingual model in challenging tasks.

As already mentioned in Section \ref{sec:introduction}, \citet{hu2020systematic} assembled a set of English syntactic tests in order to assess the syntactic generalization potential of  a number of different neural LMs (LSTM, ON-LSTM, RNNG and GPT-2). The tests are accessible through the SyntaxGym toolkit \citep{gauthier2020syntaxgym}; cf. also Section \ref{subsec:intro_syntaxgym}. Our methodology is analogous, although our objective is different. Rather than comparing the performance of several monolingual models, we contrast the performance of monolingual and multilingual transformer-based models. Furthermore, while their only test suite source is the English SyntaxGym, we create and use also a Spanish SyntaxGym; cf. Section \ref{subsec:syntaxgym_es}.

\section{Test Suites}
\label{sec:test_suites}

For the English test suites, we used SyntaxGym,\footnote{\url{http://syntaxgym.org/}} an online platform that compiles a variety of linguistic tests used by \citet{hu2020systematic} to assess the syntactic coverage of language models. It contains 34 suites, grouped into 6 different so-called \textit{circuits}, a classification based on what is required from the models to process the targeted constructions. For the Spanish test suites, we created SyntaxGymEs, adapting 11 of the existing suites for English and building 15 new ones, including a whole new circuit. In what follows, we first introduce the original English SyntaxGym and then present in detail the novel SyntaxGymEs.

\subsection{SyntaxGym for English}
\label{subsec:intro_syntaxgym}

The tests in the SyntaxGym designed by
\citet{hu2020systematic} (henceforth also referred to as ``English SyntaxGym") are based on the notion of \textit{surprisal}. A sequence of words is given to a language model, which assigns a probability to each of the following candidate words. Given the syntactic properties of the considered language, some candidate words are less surprising than others, and so should be predicted by a language model. For instance, after the sequence {\it The cat}, the inflected word {\it sleeps} should be less surprising than {\it sleep}.

Each test consists of a list of \textsc{items} that vary in a controlled way according to a set of \textsc{conditions} determined by the experimental design. The other main component is a series of \textsc{predictions} comparing surprisal values in specific regions of the items across conditions. If the relevant syntactic generalization has been learned by the model, the predictions should hold.

Moreover, some tests have versions with \textsc{modifiers}, in which additional clauses or phrases have been embedded inside each item. These modifiers increase the linear distance between two co-varying items, making the task harder. Sometimes they also include a distractor word in the middle of a syntactic dependency, which can lead the models to misinterpret the dependency.

The test suites are arranged in terms of the following {\it circuits}:

$\bullet$ \textbf{Agreement}: Morphosyntactic phenomena that occur when the features of an item constrain another item to adopt a specific form. This is a marginal phenomenon in English, so the original circuit only includes 3 test suites on \textit{Subject-verb number agreement}, all of them with modifiers \cite{marvin-linzen-2018-targeted}.

$\bullet$ \textbf{Licensing}: A construction's need for the presence of a \textit{licensor} to allow its occurrence in a sentence. The circuit consists of 4 suites on \textit{Negative polarity items} (2 of them with modifiers) and 6 on \textit{Reflexive pronouns} (all of them with modifiers), also from \citet{marvin-linzen-2018-targeted}.

$\bullet$ \textbf{Center embedding}: Subordinate clauses that sit in the middle of their superordinate clause, creating nested dependencies. This circuit contains 2 test suites: \textit{Center embedding} and \textit{Center embedding with modifier}, from \citet{wilcox2019hierarchical}.

$\bullet$ \textbf{Long-distance dependencies (LDDs)}: LDDs occur when two constituents that are syntactically related do not appear adjacent to one another, but at a longer distance from one another.
The circuit includes 6 suites on \textit{Filler-gap dependencies} (2 with modifiers and 4 addressing extraction and hierarchy) from \citet{wilcox-etal-2018-rnn} and \citet{wilcox-etal-2019-structural}, and 2 suites on \textit{Cleft structure} that were first introduced in \cite{hu2020systematic}.

$\bullet$ \textbf{Gross syntactic expectation}: Expectation for a large syntactic structure usually induced by subordinating adverbs or conjunctions.
4 test suites on \textit{Subordination} (from \citet{futrell2018rnns}, 3 of them with modifiers) constitute the circuit.

  $\bullet$ \textbf{Garden path effects}: Effects that emerge when an incorrect but locally likely parse needs to be abandoned in favor of the correct one, once a specific word appears in the sentence.
    Two such effects are considered in this circuit: \textit{Main verb/reduced relative clause (MVRR)} and \textit{NP/Z garden paths}, with respectively 2 and 4 suites, all from \citet{futrell2018rnns}.

\subsection{SyntaxGymES: SyntaxGym for Spanish}
\label{subsec:syntaxgym_es}

For Spanish, we expand the tests in \cite{hu2020systematic} so as to cover language-specific phenomena. In this section, we detail which of the original tests we retained, which ones we modified, and which ones we added within each original circuit. A whole new circuit regarding the linear order of a sentence's basic constituents was also added, since flexibility in this respect is a characteristic that distinguishes Spanish (and other Romance languages) from English. For a more detailed description with examples and predictions, see the Supplementary Material; upon acceptance of the paper, SyntaxGymES will be published in the SyntaxGym platform http://syntaxgym.org/.

\subsubsection{Notation}
We follow the usual notations in linguistic literature. An asterisk `*' preceding an example signals that the sentence is ungrammatical, it violates some principle or constraint. A question mark `?' is used to indicate a marginal sentence, i.e., a sentence that is grammatical but very uncommon or that requires a non-straightforward interpretation. The exclamation mark `!' indicates a highly difficult sentence to process for the human mind.

\subsubsection{Agreement}
Unlike English, Spanish is a morphologically rich language, and as such it presents many morpho-syntactic phenomena related to agreement. For this reason, out of the six original circuits, \textbf{Agreement} was the one that underwent the most changes.

Regarding verbal agreement (constraints imposed on the verb by the subject), we adapted two existing test suites, \textbf{Subject-Verb Agreement with Object Relative Clause} and \textbf{Subject-Verb Agreement with Subject Relative Clause}, and created a new one, \textbf{Basic Subject-Verb Agreement}, in which both person and number features were taken into consideration.

\begin{exe}
    \ex
    \gll  Tú cocinas \\
    you.2\textsc{sg}  cook.2\textsc{sg} \\
    \ex[*]
    {\gll  Tú cocinais/cocino/cocinan \\
    you.2\textsc{sg}  cook.2\textsc{pl}/1\textsc{sg}/3\textsc{pl} \\}
\end{exe}

As for nominal agreement (constraints that a noun's gender and number features can impose on the form of other words in the sentence), we also created several new test suites:
\textbf{Determinant-Noun Agreement} simply pairs a noun with the four possible forms of the definite article (\textit{el}, \textit{la}, \textit{los}, \textit{las}), while
\textbf{Adjective-Noun Agreement} pairs a noun with the four possible forms of an adjective that modifies it (we excluded articles to avoid providing extra information).

\begin{exe}
    \ex
    \gll  La tienda vende discos usados \\
    the store sells disc.\textsc{m.pl} used.\textsc{m.pl} \\
    \ex[*]
    {\gll  La tienda vende discos usados/usado/usadas/usada \\
    the store sells disc.\textsc{m.pl} used.\textsc{m.pl}/\textsc{m.sg}/\textsc{f.pl}/\textsc{f.sg} \\}
\end{exe}

In addition to these two suites, we built similar ones for \textbf{Attribute Agreement} in copulative constructions, to which we added two versions with object or subject relative clauses as modifiers, and also for \textbf{Predicative Agreement} in constructions with subject or object predicative complement. The only difference here is that the two words that must agree are not adjacent anymore. In terms of predictions, the verb/noun with matching features should have a lower surprisal than the others, and the verb/noun that matches only one feature should have a lower surprisal that the one that doesn't match any.

\subsubsection{Center Embedding}

For this circuit, we adapted to Spanish the two existing test suites in English, creating \textbf{Center Embedding} and \textbf{Center Embedding with PP modifier}.
In the basic suite, a relative clause is center embedded after the subject of the main clause. Verb transitivity and subject-verb plausibility are used to test if the models are capable of retaining the relevant information and predicting the verbs in the correct order.

\subsubsection{Gross Syntactic Expectation}

From the four original suites in this circuit, we adapted three of them: \textbf{Subordination}, and two of its versions with modifiers, \textbf{Subordination with Object Relative Clause} and \textbf{Subordination with Subject Relative Clause}.
Given a sentence that starts with a typically subordinating adverb or conjunction, these suites test the models' ability to maintain the expectation for the onset of a matrix clause for as long as the subordinate one lasts.

\subsubsection{Long-distance Dependencies}

Filler-gap dependencies are an example of LDDs. They occur when a phrase (the filler) is realized somewhere in the sentence, but is semantically interpreted at some other point (the gap).
For this circuit, we created a \textbf{Basic Filler-Gap Dependencies} test and adapted from the original English circuit a version that includes modifiers, \textbf{Filler-Gap Dependencies with Three Sentencial Embeddings}.  Embedding three sentences between filler and gap makes the task more challenging. We also adapted to Spanish the novel \textbf{Pseudo-Cleft Structures} suite introduced in \cite{hu2020systematic}.

\subsubsection{Garden Path Effects}

The Garden Path effect can be created by several syntactic ambiguities that differ cross-linguistically. The Main Verb/Reduced Relative garden path effect was the subject of two suites in the original English circuit, but it does not translate to Spanish, so those suites were not included in Spanish SyntaxGym.

On the other hand, the ambiguity responsible for  NP/Z also holds for Spanish. Here, an NP is initially interpreted as the object in a subordinate clause when it actually is the subject of the main clause (the subordinate clause having a Zero/null object). The ambiguity can be prevented with a comma, but also by placing an overt object in the subordinate clause, as is done in \textbf{NP/Z Garden Path Effect (with Overt Object)}, or by substituting its verb with a pure intransitive verb, as is done in \textbf{NP/Z Garden Path Effect (with Intransitive Verb)}. Both suites correspond to Spanish adaptations of the two original suites regarding this effect.

\begin{exe}
    \ex
    !Mientras ella leía sus manuscritos se volaron por la ventana.
    \glt !'While she read her manuscripts went out the window.'
    \ex
    Mientras ella [dormía]/[leía un libro]/[leía,] sus manuscritos se volaron por la ventana.
    \glt 'While she [slept]/[read a book]/[read,] her manuscripts went out the window.'
\end{exe}

\subsubsection{Licensing}

Negative polarity items (NPIs), like {\it any} or {\it ever} in English, are examples of words that need to be licensed by negation. Since Spanish NPIs do not function exactly in the same way, we took the original NPI Licensing test as inspiration and created two new suites: \textbf{Negative Polarity Items} and \textbf{NPIs and Polarity Agreement}.

Constructions with verbs in subjunctive mood also require the presence of a licensor. In Spanish, a verb expressing feelings (e.g. of joy, surprise, pleasantness) in the main clause, creates the expectation for subjunctive mood in the subordinate clause. This was the basis for a new test suite: \textbf{Subjunctive Mood and Verbs that Express Feeling}.

\begin{exe}
    \ex
    \gll Espero que mañana llueva/*lloverá. \\
    (I)hope that tomorrow rain.\textsc{sub}/will.rain\textsc{ind} \\
    \glt 'I hope it [rains]/[will rain] tomorrow.'
\end{exe}

The other new suite in this circuit, \textbf{Subjunctive Mood, Negation and Belief Verbs}, relies on the fact that belief verbs can also license subjunctive mood, but only when combined with negation:

\begin{exe}
    \ex
    \gll No creo que mañana llueva/*lloverá. \\
    \textsc{neg} (I)believe that tomorrow rain.\textsc{sub}/will.rain.\textsc{ind} \\
    \glt ‘I don't think it [rain]/[will rain] tomorrow.'
    \ex
    \gll Creo que mañana no lloverá/*llueva. \\
    (I)believe that tomorrow \textsc{neg} will.rain.\textsc{ind}/rain.\textsc{sub}\\
    \glt 'I think it [won't]/[don’t] rain tomorrow.'
\end{exe}

\subsubsection{Linearization}

One of the main syntactic distinctions between languages is constituent order within the sentence. But, in addition to the canonical order in which these elements appear, languages also differ in their flexibility to alter that order. Spanish allows some flexibility, which was the basis for three new test suites.

For \textbf{Subject–Auxiliary Verb–Main Verb Linearization}, the possibility to postpone the subject is compared with the rigidity of the relation between main and auxiliary verb, which must be adjacent and do not allow inversion:

\begin{exe}
    \ex
    Juan [ha comido]/*[comido ha].
    \glt 'John [has eaten]/[eaten has].'
    \ex
    Ha [comido Juan]/*[Juan comido].
    \glt *'Has [eaten John]/[John eaten].'
\end{exe}

In the \textbf{Subject–Verb–Object Linearization} test, we compare the phenomenon in affirmative versus interrogative sentences. In Spanish, word order flexibility holds for affirmative sentences, but not for interrogative ones, where subject-verb inversion is compulsory:

\begin{exe}
    \ex
    Ana compró un libro. / Compró un libro Ana.
    \glt 'Ann bought a book. / Bought a book Ann.'
    \ex
    ¿Qué compró Ana? / *¿Qué Ana compró?
    \glt 'What did Ann buy? / *What Ana did buy?'
\end{exe}

 Word order variations also appear within the NP, as captured by the \textbf{Noun-Adjective and Noun-PP Linearization} test. Contrary to English, Spanish adjectives usually come after the noun. But again, the language allows for some flexibility and they can be swapped. This possibility, however, does not apply to other noun modifiers like prepositional phrases:

\begin{exe}
    \ex
    Construyó una [mesa robusta]/[robusta mesa].
    \glt 'He built a [sturdy table]/[table sturdy].'
    \ex
    Construyó una [mesa de madera]/*[de madera mesa].
    \glt 'He built a [wooden table]/*[table wooden].'
\end{exe}

\section{Experiments}
\label{sec:experiments}

We test the base cased versions of BERT and mBERT \citep{devlin2018bert}, RoBERTa \citep{liu2019roberta} and XLM-R \citep{conneau-etal-2020-unsupervised} on the English SyntaxGym and BETO \citep{canete2020spanish}, mBERT and XLM-R on the Spanish SyntaxGym. To run the experiments, we use the SyntaxGym toolkit \citep{gauthier2020syntaxgym}.


\subsection{Experimental Setup}
\label{sec:experimental_setup}

The SyntaxGym test suites are designed from the perspective of sentence generation, i.e., with the hypothesis that if a model has correctly learned some relevant syntactic generalization, it should assign higher probability to grammatical and natural continuations of sentences. This requires asking the models to predict the next token given a context of previous tokens, in a left-to-right generative fashion. However, BERT-based and RoBERTa-based families of models (in our case, BERT and mBERT on the one side, and RoBERTa and XLM-R on the other side) are bidirectional, they are trained with a masked language modeling objective to predict a word given its left and right context. In this work, we follow \citet{wang2019bert}'s sequential sampling procedure to evaluate the probability of a text sequence, encoding unidirectional context in the forward direction. To compute the probability distribution for a sentence with $N$ tokens, we start with a sequence of $N+2$ tokens: a \textit{begin\_of\_sentence} token plus $N+1$ \textit{mask} tokens, where the last \textit{mask} corresponds to the \textit{end\_of\_sentence} token. For each token position \textit{i} in $[1, N]$, we compute the probability distribution over the vocabulary given the left context of the original sequence, and select the probability assigned by the model to the original word.

For example, in an agreement test with the sentence `\textit{The girls run fast.}', a model that has properly learned agreement should assign a higher probability to \textit{run} than to \textit{runs} for the third word. In order to test it, we feed the tokens sequence \textit{[[bos] [The] [girls] [mask] [mask] [mask] [mask]]} to the model, and compare the probabilities assigned by the model to \textit{run} and \textit{runs} for position 4.

\subsection{Results of the experiments}
\label{sec:results}

This section summarizes the results of our experiments that aim to: (i) contrast the performance of monolingual and multilingual models on English and Spanish and (ii) provide insights on the performance of the multilingual models across languages.

\begin{figure}[t]
    \includegraphics[width=0.49\textwidth]{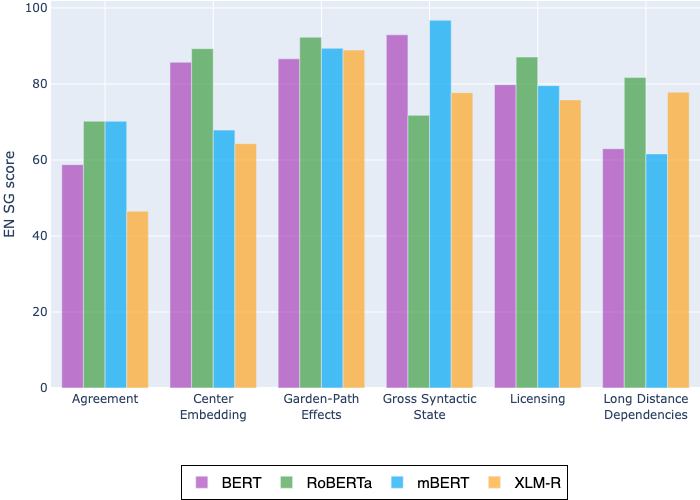}
    \caption{Performance accuracy across English circuits}
    \label{fig:models_performance_across_tags_multilingual_en_bars}
\end{figure}

\begin{figure}[h]
    \includegraphics[width=0.49\textwidth]{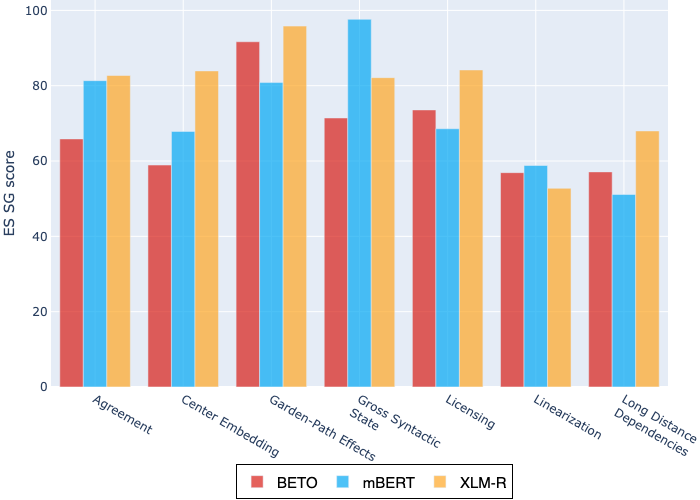}
    \caption{Performance accuracy across Spanish circuits}
    \label{fig:models_performance_across_tags_multilingual_es_bars}
\end{figure}

Table \ref{tab:sg_performance} shows the average SyntaxGym (SG) performance of the evaluated monolingual and multilingual models on the English and Spanish SyntaxGyms. Figures \ref{fig:models_performance_across_tags_multilingual_en_bars} and \ref{fig:models_performance_across_tags_multilingual_es_bars} zoom in on the performance of the tested models with respect to specific circuits for English and Spanish respectively.

\begin{table}
    \centering
    \begin{tabular}{lcc}
    \hline
      &   \multicolumn{2}{c}{\textbf{Average SG performance}} \\
    \hline
    \textbf{Model}  &   \textbf{English}   &   \textbf{Spanish} \\
    \hline
    BERT    &   77.80   &   --- \\
    RoBERTa &   82.04   &   --- \\
    mBERT   &   77.55   &   72.31 \\
    XLM-R   &   71.84   &   78.50 \\
    BETO    &   ---     &   67.92 \\
    \hline
    \end{tabular}
    \caption{Average SG score by model class for the English and Spanish tests.}\label{tab:sg_performance}
\end{table}

\begin{figure}[t]
    \includegraphics[width=0.49\textwidth]{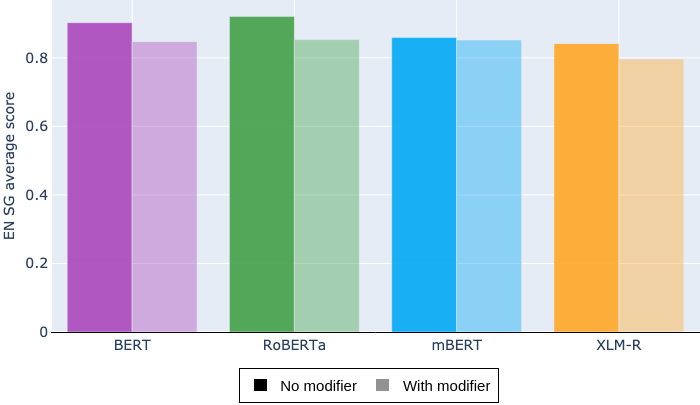}
    \caption{Models average English SG score in Center Embedding, Cleft structure, MVRR, NPZ-Verb, NPZ-Object and Subordination, with and without modifiers.}
    \label{fig:sg_score_modifiers_multilingual_en}
\end{figure}

\begin{figure}[t]
    \includegraphics[width=0.49\textwidth]{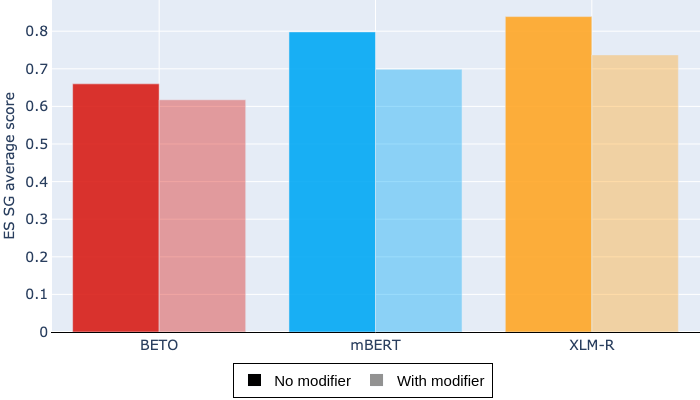}
    \caption{Models average Spanish SG score in Attribute Agreement, Subject-Verb Agreement, Subordination, Center Embedding and Filler-Gap Dependencies, with and without modifiers.}
    \label{fig:sg_score_modifiers_multilingual_es}
\end{figure}

Six of the English test suites (Center Embedding, Cleft structure, MVRR, NPZ-Verb, NPZ-Object, Subordination) and five of the Spanish test suites (Attribute Agreement, Basic  Subject-Verb Agreement, Subordination, Center Embedding, Basic Filler-Gap Dependencies) include tests with and without modifiers, i.e,. intervening content inserted before the critical region. Figures \ref{fig:sg_score_modifiers_multilingual_en} and \ref{fig:sg_score_modifiers_multilingual_es} show the models' average scores in these test suites, without modifiers (dark bars) and with modifiers (light bars), evaluating how robust each model is with respect to the corresponding content.

\section{Discussion}
\label{sec:discussion}

Let us assess in detail the results of the experiments from above. In what follows, we compare the performance of monolingual with the performance of multilingual models and analyze the cross-language performance of multilingual models, as well as the stability of the individual models with respect to modifiers.

\subsection{Monolingual vs multilingual models}

RoBERTa shows an overall higher performance than the other models for English (Table \ref{tab:sg_performance}). This is not surprising since it is trained on 10 times more data than BERT, and it has been shown to improve over BERT in many NLU tasks. However, while mBERT does not seem to lose performance compared to BERT, XLM-R loses around 10 points compared to RoBERTa. As XLM-R is specifically designed to offer a more balanced performance across languages, with a special focus on low-resource languages, it appears natural that it loses some performance on high-resource languages such as English. For Spanish, the multilingual models clearly outperform the monolingual model. This is likely due to the fact that while BETO and mBERT are of comparable size and are trained with the same amount of data (16GB), BETO is only trained with a Masked Language Modeling (MLM) objective, and mBERT is trained on MLM and Next Sentence Prediction (NSP). On the other hand, XLM-R is also only trained on MLM, but it is trained on more than 2TB of data, 53 GB corresponding to Spanish data.

RoBERTa outperforms all other models in all the English circuits (cf. Figure \ref{fig:models_performance_across_tags_multilingual_en_bars}), except in Gross Syntactic State, in which BERT-based models clearly outperform RoBERTa-based models, and the multilingual model outperforms the monolingual one in both families. Intuitively, we believe that the NSP training objective of BERT-based models helps them to better understand the relation between two sentences, and this knowledge can also be applied to the relation between two clauses (which is the basis of the Gross Syntactic State circuit). Comparing the BERT and RoBERTa model families, it is interesting to notice that while RoBERTa outperforms XLM-R in all circuits except Gross Syntactic State, BERT only outperforms mBERT in 3 of them.

Interestingly, all models seem to struggle with Agreement in English. This observation is aligned with \citet{mueller2020cross}'s hypothesis that language models learn better hierarchical syntactic generalizations in morphologically complex languages (such as, e.g., Spanish), which frequently provide  overt cues to syntactic structure, than in morphologically simpler languages (such as, e.g., English). Indeed, the fact that XLM-R offers the lowest performance may be related to the fact that the model has been more exposed to more complex languages than the others. For Long Distance Dependencies, BERT-based models show a low performance compared to RoBERTa-based models. This might be due to the different training procedures adopted in both model families (i.e., that RoBERTa does not include the Next Sentence Prediction task (as BERT does) and introduces dynamic masking).


On the other hand, in specific circuits for Spanish (cf. Figure \ref{fig:models_performance_across_tags_multilingual_es_bars}) XLM-R outperforms the other two models in 5 out of 7 circuits. As observed for English, the BERT-based models struggle with the Long Distance Dependencies tests, and mBERT offers an outstanding performance in Gross Syntactic State. The monolingual model, BETO, is outperformed by mBERT in 4 out of 7 tests, and by XLM-R in all 6 out of 7 tests. As mentioned before, these differences may be related to the fact that, unlike BERT, BETO is not trained with the NSP objective; but also to the difference in training data size: 16GB for BETO vs. more than 2TB (of which 53GB of Spanish data) for XLM-R.

All models offer a low performance in the new Linearization test for Spanish. A more in-depth investigation is necessary to explain this. The test has been designed with literary Peninsular Spanish in mind, and it is possible that the training data may not contain enough samples that show the targeted word order varieties, or may contain data from American Spanish sources, which may show differences in canonical word order with respect to Peninsular Spanish.

\subsection{Cross-language multilingual models performance}

As shown in Table \ref{tab:sg_performance}, multilingual models do not syntactically generalize equally well in both languages. While mBERT offers a better generalization in English, outperforming XLM-R by almost 6 points, XLM-R generalizes better in Spanish, outperforming mBERT by 6 points. This observation corroborates our intuition that XLM-R sacrifices performance in high-resource languages (e.g., English, with 300GB of training data) to be able to offer a more balanced performance across languages (e.g., Spanish, with 53GB of training data).

Comparing Figures \ref{fig:models_performance_across_tags_multilingual_en_bars} and \ref{fig:models_performance_across_tags_multilingual_es_bars}, we observe improvements in the Spanish tests for XLM-R in 4 out of 6 circuits, particularly noticeable in Agreement and Center Embedding, while it loses around 10 points in Long Distance Dependencies. On the other hand, mBERT also shows a big improvement in the Spanish tests in Agreement, while it loses performance in Garden Path Effects, Licensing and Long Distance Dependencies.

\subsection{Model stability with respect to modifiers}

Since modifiers increase the linear distance between the elements in a dependency structure, thus making the task more demanding, stability in this respect indicates that models have robustly learnt the appropriate syntactic generalization and do not depend that much on adjacency. Figures \ref{fig:sg_score_modifiers_multilingual_en} and \ref{fig:sg_score_modifiers_multilingual_es} show the models' average scores in those test suites that have two versions: without modifiers (dark bars) and with modifiers (light bars). As was intuitively expected, all the models offer a higher performance in the tests without modifiers. While for English the multilingual models are the less affected, for Spanish BETO seems to be more robust than the multilingual models, even though it offers a lower performance.

\section{Conclusions}
\label{sec:conclusions}

In this paper, we assessed the syntactic generalization potential of selected transformer-based language models on English and Spanish. We have shown that multilingual models do not generalize equally well across languages: mBERT generalizes better for phenomena in English, while XLM-R does it better for phenomena in Spanish.
We have also shown that the answer to the question whether monolingual or multilingual models generalize better is equally
language-specific: the monolingual RoBERTa generalizes better on English, while the multilingual XLM-R generalizes better on Spanish. While it is possible that the multilingual abstractions captured by XLM-R become useful for morphologically rich languages such as Spanish, this difference may also be related to the difference in the amount of training data used to train BETO and XLM-R, and therefore it is possible that a monolingual model trained with a comparable amount of data could outperform the multilingual models.

The performance of all models is affected by the presence of modifiers, which shows that the complexity of the syntactic structure is still a challenge. In general, each syntactic phenomenon deserves attention. For instance, Agreement in English is hard to learn, given the scarcity of cues (especially if compared to a morphologically rich language), and so is Linearization in Spanish.

As far as the nature of the training procedures of the models is concerned, the lack of Next Sentence Prediction (NSP) objective in the RoBERTa model family seems to harm BETO, but not XLM-R; this suggests that the performance of BETO may be improved with (much) more training data. It also seems to harm in the case of the Gross Syntactic State circuit, suggesting that RoBERTa-based models may also benefit from complementary training objectives in their pretraining procedure.

Overall, our experiments have also shown the importance of testing models on a wider range of languages, in particular, morphologically rich ones. As part of our future work, we plan to expand further SyntaxGymES and develop SyntaxGyms for a number of other selected languages. Also, careful examination of a wider range of material is necessary to ensure that important phenomena are not left out, so as to assess the actual coverage of the test suites.

\section*{Acknowledgments}
This work has been partially funded by the European Commission via its H2020 Research Program under the contract numbers 779962, 786731, 825079, and 870930.

\bibliographystyle{acl_natbib}
\bibliography{anthology,acl2021}

\appendix

\section{Spanish SyntaxGym: Description of Test Suites}

This appendix lists and describes all the test suites compiled for Spanish SyntaxGym. Each test consists of a list of \textsc{items} that vary in a controlled way according to a set of \textsc{conditions} determined by the experimental design. A series of \textsc{predictions} compare surprisal values at specific regions of the items across conditions. Some tests have versions with \textsc{modifiers} that increase the linear distance between two co-varying items, making the task more demanding.

The test suites are arranged in terms of \textit{circuits} of related syntactic phenomena. Each of the following sections corresponds to one of these circuits.
\vspace{0.2cm}

\textbf{Notation}. An asterisk * signals an ungrammatical sentence, a question mark ? indicates a marginal sentence (grammatical but very uncommon or requiring a difficult interpretation), an exclamation point ! denotes high processing difficulty.

\subsection{Agreement}
Agreement is a morpho-syntactic phenomenon that occurs when the features of an item constrain another item to adopt a specific form.

\vspace{0.2cm}
$\bullet$ \textbf{Basic Subject-Verb Agreement}.
New suite. Spanish finite verbs in any tense/mood have six inflected forms according to person and number features. The verb's features the subject's, otherwise the result is ungrammatical.
\vspace{-0.2cm}
\begin{exe}
    \ex
    \gll  Tú cocinas \\
    you.2\textsc{sg}  cook.2\textsc{sg} \\
    \vspace{-0.2cm}
    \ex[*]
    {\gll  Tú cocinais/cocino/cocinan \\
    you.2\textsc{sg}  cook.2\textsc{pl}/1\textsc{sg}/3\textsc{pl} \\}
\end{exe}
\underline{Predictions}: The surprisal at the verb region is expected to be lower when it matches the subject than in any other condition. It is also expected to be lower when at least one of the features (person or number) agrees than when both disagree.

\vspace{0.2cm}
$\bullet$ \textbf{Subject-Verb Agreement with Subject Relative Clause}.
Adapted from English. This test focuses on number agreement. The subject relative clause includes a \textit{distractor} NP differing in number with the subject.
\vspace{-0.2cm}
\begin{exe}
    \ex
    \gll  El fontanero que ayudó a los albañiles trabaja/*trabajan los sábados. \\
    the.\textsc{sg} plumber that helped.3\textsc{sg} to the\textsc{pl} bricklayers work.3\textsc{sg}/3\textsc{pl} the saturdays. \\
    \glt 'The plumber who helped the bricklayers works/*work on saturdays.'
    \vspace{-0.2cm}
    \ex
    \gll  Los fontaneros que ayudaron al albañil *trabaja/trabajan los sábados. \\
    the.\textsc{pl} plumbers that helped.3\textsc{sg} to.the\textsc{pl} bricklayer work.3\textsc{pl}/3\textsc{sg} the saturdays. \\
    \glt 'The plumbers who helped the bricklayer *works/work on saturdays.'
\end{exe}
\vspace{-0.2cm}
\underline{Predictions}: A successful model should place higher probability to the verb agreeing with the subject (instead of the distractor) both in singular and in plural.

\vspace{0.2cm}
$\bullet$ \textbf{Subject-Verb Agreement with Object Relative Clause}.
Adapted from English. Equal to the previous one, but with an object relative clause.

\vspace{0.2cm}
Nominal agreement was the basis for the following 6 new test suites. All of them share the same predictions: the surprisals should be lower when both gender and number features in the second word of the agreement relation match those in the first word. They should also be lower when only one of the features agrees than when both disagree.

\vspace{0.2cm}
$\bullet$ \textbf{Determiner-Noun Agreement}.
New suite. The four possible forms of the definite article are paired with different nouns.
\vspace{-0.2cm}
\begin{exe}
    \ex
    \gll  El/*La/*Los/*Las gato  \\
    the.\textsc{m.sg/*f.sg/*m.pl/*f.pl} cat \\
\end{exe}

$\bullet$ \textbf{Adjective-Noun Agreement}.
New suite. The test pairs a noun with the four possible forms of an adjective that modifies it (we used constructions without determiner to avoid providing the models with extra information).
\vspace{-0.1cm}
\begin{exe}
    \ex
    \gll  La tienda vende discos usados/*usado/*usadas/*usada \\
    the store sells discs used.\textsc{m.pl}/\textsc{m.sg}/\textsc{f.pl}/\textsc{f.sg} \\
    \glt 'The store sells second-hand discs.'
\end{exe}

$\bullet$ \textbf{Attribute Agreement}.
New suite. Here, a noun is paired with and adjective through a copulative construction. This suite has 2 versions with object or subject relative clauses as modifiers.
\vspace{-0.1cm}
\begin{exe}
    \ex
    \gll  El piso está vacío/*vacía/*vacíos/*vacías\\
    the flat is empty.\textsc{m.sg}/*\textsc{f.sg}/*\textsc{m.pl}/*\textsc{f.pl}\\
\end{exe}

$\bullet$ \textbf{Predicative Agreement}.
New suite. The subject or the object is paired with an adjective functioning as a predicative complement.
\vspace{-0.1cm}
\begin{exe}
    \ex
    \gll  Los niños llegaron cansados/*cansado/*cansadas/*cansada\\
    the children arrived tired.\textsc{m.pl}/*\textsc{m.sg}/*\textsc{f.pl}/*\textsc{f.sg} \\
    \glt 'The children arrived tired.'
\end{exe}

\subsection{Center Embedding}
A center embedded clause is a subordinate clause that sits in the middle of its superordinate clause, creating nested dependencies that may be challenging for the models.

\vspace{0.2cm}
$\bullet$ \textbf{Center Embedding}.
Adapted from English. A relative clause is center embedded after the subject of the main clause. Verb transitivity and subject-verb plausibility are used to test if the models are capable of retaining the relevant information and predicting the verbs in the correct order.
\vspace{-0.1cm}
\begin{exe}
    \ex
    La tormenta que el capitán [capeó amainó]/?[amainó capeó].
    \glt 'The storm the captain [weathered abated]/?[abated weathered].'
\end{exe}
\vspace{-0.2cm}
\underline{Prediction}: The surprisal of the combination of verbs should be smaller when their relative order creates a plausible sentence than when it creates an implausible one.

\vspace{0.2cm}
$\bullet$ \textbf{Center Embedding with modifier}.
In the version with modifier, a prepositional phrase is inserted after the subject of the subordinate clause.

\subsection{Gross Syntactic State}
Expectation for a  large  syntactic  structure  at  some  point  within the sentence.

\vspace{0.2cm}
$\bullet$ \textbf{Subordination}. Adapted from English. A sentence starting with a subordinate clause creates the expectation for the onset of a matrix clause for as long as the subordinate one lasts.
\vspace{-0.1cm}
\begin{exe}
    \ex
    ?(Mientras) ella miraba los resultados, el doctor entró en la habitación.
    \glt 'While she looked at the results, the doctor entered the room.'
    \ex
    (*Mientras) ella miraba los resultados.
    \glt '(*While) she looked at the results.'
\end{exe}
\vspace{-0.1cm}
\underline{Predictions}: The surprisal for the lack of a second clause should be higher when there is a subordinating conjunction or adverb than where there is not. But having two clauses joined by a conjunction/adverb should be less surprising than their juxtaposition.

\vspace{0.2cm}
$\bullet$ \textbf{Subordination with Object Relative Clause} and \textbf{Subordination with Subject Relative Clause}. Adapted from English. Versions of the previous suite but with a modifier.

\subsection{Long-distance Dependencies}

LDDs occur when two syntactically related groups do not appear adjacent to one another but at a longer distance from one another.

\vspace{0.2cm}
$\bullet$ \textbf{Basic Filler-Gap Dependencies}. New suite, a simplified version of the existing FGD tests for English. FGDs occur when a phrase (the filler) is realized somewhere in the sentence but is semantically interpreted at some other point (the gap).

\begin{exe}
    \ex
    Yo sé [lo que]/*que tu amigo tiró \_ al suelo.
    \glt 'I know what/*that your friend threw \_.'
    \ex
    Yo sé *[lo que]/que tu amigo tiró una colilla al suelo.
    \glt 'I know *what/that your friend threw a cigarette butt.'
\end{exe}
\vspace{-0.1cm}
\underline{Predictions}: The overt object should be more surprising when there is a filler when there is not. We also expect lower surprisal when the sentence has a filler later followed by gap than when it has a conjunction instead but the gap remains.

\vspace{0.2cm}
$\bullet$ \textbf{Filler-Gap Dependencies with Three Sentencial Embeddings}. Adapted from English. It is a version of the previous test that includes a modifier (three sentential embeddings) between filler and gap. This makes the task more challenging. The predictions, though, remain the same.

\vspace{0.2cm}
$\bullet$ \textbf{Pseudo-Cleft Structures} Adapted from English. A pseudo-cleft or wh-cleft is formed by a wh-element extracting content from a relative clause joined by a copula to a constituent that provides the content requested by the wh-element. The extracted constituent can be a NP or a VP. In the VP case, the verb in the relative clause must be an inflected form of ‘hacer’ (‘to do’).
\vspace{-0.1cm}
\begin{exe}
    \ex
    Lo que tú difundiste/?hiciste fue un rumor.
    \glt 'What you spread/*did was a rumor.'
    \ex
    Lo que tú *difundiste/hiciste fue confirmar un rumor.
    \glt 'What you *spread/did was confirm a rumor.'
\end{exe}
\vspace{-0.1cm}
\underline{Predictions}: The surprisal should be lower for the extracted VP when the verb in the relative clause is a light verb (\textit{hacer} – ‘to do’) than when it is not, but it should be higher for the extracted NP when the verb is light than when it is semantically heavier and matches the NP. In addition, the difference in the first case should be more important than in the second one. This happens because the light verb admits a wider range of objects, whereas in the first case, one of the options is syntactically incorrect.

\subsection{Garden Path Effects}

Garden-path effects emerge when an incorrect but locally likely parse needs to be abandoned in favor of the correct one. In the NP/Z garden path, an NP is initially interpreted as the object in a subordinate clause, but when the main verb appears, this NP should be reinterpreted as its subject. The effect can be prevented by adding a comma, but also by placing an overt object in the subordinate clause, or by substituting its verb with a purely intransitive one. These are the basis for the next two suites.

\vspace{0.2cm}
$\bullet$ \textbf{NP/Z Garden Path Effect (Overt Object)}.

$\bullet$ \textbf{NP/Z Garden Path Effect (Intransitive Verb)}. Both adapted from English.
\vspace{-0.1cm}
\begin{exe}
    \ex
    !Mientras ella leía sus manuscritos se volaron por la ventana.
    \glt !'While she read her manuscripts went out the window.'
    \ex
    Mientras ella [dormía]/[leía un libro]/[leía,] sus manuscritos se volaron por la ventana.
    \glt 'While she [slept]/[read a book]/[read,] her manuscripts went out the window.'
\end{exe}
\vspace{-0.1cm}
\underline{Predictions}: The main verb should be more surprising in the garden path condition than when the effect has been prevented either by the comma or by interfering with the verb. Moreover, the difference in surprisal should be bigger when the comma is essential to solve the garden path effect than when it is not.

\subsection{Licensing}

In natural language, some words or constructions need the presence of a licensor to allow their occurrence in a sentence. This happens with NPIs (Negative polarity items) and subjunctive mood, for instance.

\vspace{0.2cm}
$\bullet$ \textbf{Negative Polarity Items and Polarity Agreement}. New suite. In Spanish, NPIs that follow the verb (such as \textit{nunca} 'never', \textit{nadie} 'nobody’, and \textit{nada} 'nothing’) need to be licensed by negation. This ‘double negative’ does not result in an affirmative, it is a sort of polarity agreement.
\vspace{-0.1cm}
\begin{exe}
    \ex
    \gll Yo no bebo nunca/?siempre. \\
    I \textsc{neg} drink never/always \\
    \glt 'I never drink./I don't drink always.'
    \ex
    Yo bebo *nunca/siempre.
    \glt 'I *ever/always drink.'
\end{exe}
\vspace{-0.1cm}
\underline{Predictions}: We expect the surprisals in both agreeing conditions (negative-NPI, positive-PPI) to be lower than in any of the non-agreeing conditions (negative-PPI, positive-NPI).

\vspace{0.2cm}
$\bullet$ \textbf{Negative Polarity Items}. New suite. NPIs also need to be in the scope of the negation to be licensed by it. This suite compares between a negative particle that ``commands'' the NPI and one that doesn’t.
\vspace{-0.1cm}
\begin{exe}
    \ex
    \gll Tú, como no mirabas por la ventana, *(no) has visto a nadie. \\
    You, as \textsc{neg} looked by the window, \textsc{neg} have seen at nobody \\
    \glt 'As you weren't looking through the window, you have *(not) seen anybody.'
    \ex
    \gll Tú, como mirabas por la ventana, *(no) has visto a nadie. \\
    You, as looked by the window, \textsc{neg} have seen at nobody \\
    \glt 'As you were looking through the window, you have *(not) seen anybody.'
\end{exe}
\vspace{-0.1cm}
\underline{Predictions}: The NPI should be more surprising when there isn't a negative particle that commands it, independently of the presence of another one that does not command it. 

\vspace{0.2cm}
$\bullet$ \textbf{Subjunctive Mood and Verbs that Express Feeling}. New suite. Feeling verbs that introduce a subordinate clause serve as licensors for subjunctive mood, whereas other type of verbs do not.
\vspace{-0.1cm}
\begin{exe}
    \ex
    \gll Espero que mañana llueva/*lloverá. \\
    (I)hope that tomorrow rain.\textsc{sub}/will.rain.\textsc{ind} \\
    \glt 'I hope it rains/*[will rain] tomorrow.'
    \ex
    \gll Sé que mañana *llueva/lloverá. \\
    (I)know that tomorrow rain.\textsc{sub}/will.rain.\textsc{ind}\\
    \glt 'I know it [will rain]/rains tomorrow.'
\end{exe}
\vspace{-0.1cm}
\underline{Predictions}: Subjunctive mood should be less surprising than indicative mood when the verb in the main clause expresses feelings. But when it doesn't, subjunctive should be more surprising than indicative mood. Moreover, subjunctive mood should also be more surprising with a feeling verb than with a non-feeling verb.

\vspace{0.2cm}
$\bullet$ \textbf{Subjunctive Mood, Negation and Belief Verbs}. New suite. Belief verbs can also license subjunctive mood, but only when combined with negation.
\vspace{-0.1cm}
\begin{exe}
    \ex
    \gll No creo que mañana llueva/*lloverá. \\
    \textsc{neg} believe that tomorrow rain.\textsc{sub}/will.rain.\textsc{ind} \\
    \glt 'I don't think it rains/[will rain] tomorrow.'
    \ex
    \gll Creo que mañana no *llueva/lloverá. \\
    (I)believe that tomorrow \textsc{neg} rain.\textsc{sub}/will.rain.\textsc{ind}\\
    \glt 'I think it rains/[won't rain] tomorrow.'
\end{exe}
\vspace{-0.1cm}
\underline{Predictions}: The subordinate verb should be less surprising in subjunctive than in indicative mood when the main clause is negated. However, the contrary should hold when the subordinate clause is negated but the main one is not. In addition, subjunctive mood should be less surprising when the negation is in the main clause than when it is in the subordinate clause.

\subsection{Linearization}

Constituent order is commonly used in linguistics as a way to classify languages. But, in addition to the canonical order in which elements appear, languages also differ in their flexibility to alter that order.

\vspace{0.2cm}
$\bullet$ \textbf{Subject – Auxiliary Verb – Main Verb Linearization}. New suite. Subject-verb order admits inversion in Spanish but main and auxiliary verb do not and they must be adjacent.
\vspace{-0.1cm}
\begin{exe}
    \ex
    Juan ha comido. / Ha comido Juan
    \glt 'John has eaten. / Has eaten John.'
    \ex
    *Juan comido ha. / *Ha Juan comido.
    \glt 'John eaten has. / Has John eaten.'
\end{exe}
\vspace{-0.1cm}
\underline{Predictions}: The postposed subject should be less surprising than any of the alterations involving auxiliary and main verb. The canonical SV order, however, should be less surprising than postposing the subject, and the difference in this case should be less important than the differences in the first two cases.

\vspace{0.2cm}
$\bullet$ \textbf{Subject – Verb – Object Linearization}. New test. In Spanish, word order flexibility holds for affirmative sentences but not for interrogative ones, where subject-verb inversion is compulsory.
\vspace{-0.1cm}
\begin{exe}
    \ex
    Ana compró un libro/Compró un libro Ana.
    \glt 'Ann bought a book. / Bought a book Ann.'
    \ex
    ¿Qué compró Ana? / ¿Qué Ana compró?
    \glt 'What did Ana buy? / 'What Ana did buy?'
\end{exe}
\vspace{-0.1cm}
\underline{Predictions}: A postposed subject in an affirmative sentence should be less surprising than lack of SV inversion in an interrogative one. The canonical SV order in the affirmative sentence, however, should be less surprising than postposing the subject, and the difference in this case should be less important than the difference in the first one.

\vspace{0.2cm}
$\bullet$ \textbf{Noun-Adjective and Noun-PP Linearization}. New suite. Spanish adjectives usually come after the noun, but this order can be inverted. Other noun modifiers like prepositional phrases cannot.
\vspace{-0.1cm}
\begin{exe}
    \ex
    Construyó una [mesa robusta]/[robusta mesa].
    \glt 'He built a [sturdy table]/[table sturdy].'
    \ex
    Construyó una [mesa de madera]/*[de madera mesa].
    \glt 'He built a [wooden table]/*[table wooden].'
\end{exe}
\vspace{-0.1cm}
\underline{Predictions}: A PP preceding the noun should be more surprising than one following it. An adjective preceding the noun should also be more surprising than one following it, but the difference in this case should be less important than in the first one.

\end{document}